\documentclass{article}
\usepackage{spconf,amsmath,graphicx}
\usepackage{adjustbox}
\usepackage{multicol, blindtext}

\title{SPRING-INX: A Multilingual Indian Language Speech Corpus by SPRING Lab, IIT Madras}
\name{\begin{tabular}{c} Nithya R*, Malavika S*, Jordan F*, Arjun Gangwar*, Metilda N J*, S Umesh*  Rithik Sarab** \\ Akhilesh Kumar Dubey**, Govind Divakaran**, Samudra Vijaya K**, Suryakanth V Gangashetty** \end{tabular}}
\address{* SPRING Lab (formerly Speech Lab), Indian Institute of Technology Madras, India \\ ** K L University, India}

\begin{document}
%
\maketitle
\begin{abstract}
India is home to a multitude of languages of which 22 languages are recognised by the Indian Constitution as official. Building speech based applications for the Indian population is a difficult problem owing to limited data and the number of languages and accents to accommodate. To encourage the language technology community to build speech based applications in Indian languages, we are open sourcing SPRING-INX data which has about 2000 hours of legally sourced and manually transcribed speech data for ASR system building in Assamese, Bengali, Gujarati, Hindi, Kannada, Malayalam, Marathi, Odia, Punjabi and Tamil. This endeavor is by SPRING Lab , Indian Institute of Technology Madras and is a part of National Language Translation Mission (NLTM), funded by the Indian Ministry of Electronics and Information Technology (MeitY), Government of India. We describe the data collection and data cleaning process along with the data statistics in this paper.

\end{abstract}
\begin{keywords}
SPRING-INX, NLTM, manually transcribed
\end{keywords}
\section{Introduction}
As the Republic of India is characterized by its multilingual population, there are several issues faced by people moving to a different part of India. Also, there are scenarios like Parliament sessions, Television debates, interviews etc. where people require translators to communicate with others due to language barriers. To overcome the language barrier, National Language Translation Mission (NLTM) was initiated by the Government of India, and is funded by the Indian Ministry of Electronics and Information Technology (MeitY). The objectives of NLTM are:
\begin{itemize}
    \item To build Speech-to-Speech Machine Translation systems for major Indian languages
    \item To create an ecosystem involving startups and academia to build and deploy services in Indian languages
    \item To increase the internet content of Indian Languages in different domains
\end{itemize}

As part of the Speech Consortium of the NLTM-R\&D which is led by Indian Institute of Technology Madras (IITM), SPRING Lab of  IITM has collected and is collecting legally sourced and manually transcribed speech corpus in various Indian languages such as Tamil, Hindi, Indian English, Marathi, Bengali, Malayalam, Telugu, Assamese, Kannada, Gujarati, Odia, Punjabi. Bodo and Manipuri through speech data collection agencies identified using a tendering process. The data collected has been carefully evaluated by the Speech Quality Control (SQC) team led by KL University.
	
We are releasing the first set of valuable data amounting to 2000 hours (both Audio and corresponding manually transcribed transcriptions) which was collected, cleaned and prepared for ASR system building in 10 Indian languages such as Assamese, Bengali, Gujarati, Hindi, Kannada, Malayalam, Marathi, Odia, Punjabi and Tamil in the public domain. We are planning to release the remaining data subsequently in the future. We are also releasing the ESPnet recipes so that interested people can build Automatic Speech Recognition Systems. We are also planning to release meta data in the near future so that the data will be useful for building other speech-based systems such as Speaker Recognition, Speaker Diarization, Language Diarization, Language Recognition, Keyword Spotting etc.

\section{Data Collection Guidelines}
IIT Madras as part of the Speech Consortium of NLTM identified speech data collection agencies who can collect good quality speech data and can provide corresponding good quality manually transcribed data through a tendering process. It was decided to collect data from mock conversations as well as monologues on different topics. IIT Madras and the speech consortium members together prepared the guidelines for speech data collection so that the format is consistent across the vendors. The guidelines are presented below for the benefit of others who may be collecting speech data.
	
 \begin{itemize}
 \item \textbf{Speaker-related:}
 \begin{itemize}
    \item The data should be gender balanced and allowable tolerance is 10 \% 
    \item Age of the speakers must be between 18 and 60 years.
 \end{itemize}
 \item \textbf{Dialect and Domain-related:}
 \begin{itemize}
    \item At least 4 dialects per language must be collected.
 
 \item The data must be from different domains. Example domains are: weather, different types of news, entertainment, health, agriculture, education, jobs, BPO (Business Process Outsourcing) etc.
 \item Data can be a mockup or it could be actual data from agencies willing to share with proper permissions
 \item No profanity words should be used in the recording.
 \item Speech content which is communally or politically or racially biased should be avoided.  
 \end{itemize}
 \item \textbf{Audio-related:}
\begin{itemize}
 \item All audio files must be in .wav format and the sampling frequency for wide-band must be 16 kHz and for narrow-band it can either be 8 kHz or 16 kHz.
 \item Audio Coding Scheme should be 16-bit Linear PCM and the channel type should be Mono
 \item Audio Should not be post-processed or pre-processed (e.g. compression, reverb, normalization).
 \item All audio files must be intelligible with low background noise, less distortion and no audio clip-off.
\end{itemize}
 \item \textbf{Transcription-related:}
 \begin{itemize}
      \item Correct audio segmentation for transcription.
 \item Correct Transcription as per the audio.
 \end{itemize}

 \item \textbf{Speech Mode-related:}
 
 Monologues and conversational speech must be collected based on the following guidelines:
 \begin{itemize}
      \item Monologue (Single Speaker Speech) : Could be read or extempore speech.
      \item Conversational Speech : 2 to 4 speakers chatting about a topic of interest (for example: sports, news, weather, entertainment, politics, business, everyday problems like public transport, e-governance, government schemes like aadhar, dhan yojana, ayush bharat etc). Speakers should prepare a little before the conversation to avoid situations like “you tell”, “what’s up”, “okay” “what shall i say” and no real conversation. Each conversation could last about 3-5mins between different speakers.
      \item The number of males and females should be more or less equal in multiple sessions. 
      \item The conversations can be made across various smartphones using voice/data calls.
      \item Most of the data must be monolingual in nature.
\item About 10\% of conversation speech data should be both code mixed and code switched. We prefer if it is a mix of English, and another local language. (e.g., Tamil+English (Tanglish) or Hindi+English (Hinglish) etc.)

\item It is expected to provide 2 – 3 seconds of silence before starting the actual speech.
\item At least 10 different domains must be chosen.
\item Min 10 mins of data per speaker and max 30 mins per speaker. In the overall corpus, no speaker should have more than 30 minutes of audio.
\item Proper Segments file must accompany the recording. Segments are important to indicate the structural boundaries of an audio file, such as the type of sound, turns, corresponding transcriptions. 
 \end{itemize}

 \item{\textbf{Metadata-related:}}
 \begin{itemize}
     \item For both Monologue and Conversational speech data identity of speakers must be protected but should be uniquely mapped throughout the project
 \item Type of speech : Monologue or Conversation must be indicated
 \end{itemize}

 \end{itemize}
                
\section{Data Quality Assessment}
Once the data was handed over by the vendors, quality checks were performed by KL University and SPRING Lab members. Different quality checks performed are listed below:
\begin{itemize}
    \item Manually check if the data collected follows the specified formats for both audios and texts (json files, wav files).
    \item Manually check if the data has speaker information, type of data mentioned (Monologue, conversational) etc.
    \item Manually check if the text in the transcription matches the text spoken in the audio for random samples. Automatic CTC alignment based audio and transcription matching was also assessed for languages with already existing ASR systems.
    \item Manually check if the audio data is noisy for random samples.
\end{itemize}

The SQC team will check and provide detailed feedback in order to accept or reject a particular batch of recordings and transcription.

\section{Data Cleaning}
\subsection{Speech data cleaning:}
Initially we CTC aligned the audio and transcriptiom to check if the text in the transcription matches the text spoken in the audio and eliminated the speech utterances that didn’t match.
We also upsampled the audios that were in frequency 8KHz to 16KHz so that it is constant across the data. Interested people can contact us for the data in original 8kHz data for building narrow band models.

\subsection{Transcription cleaning:}
We have removed unwanted whitespace characters that were present in the transcriptions. We have also removed non-printable UTF-8 characters that were present in the transcriptions. Secondly we have removed the tags in the transcription as the tags were not necessary for ASR system building. We can also provide text in original form before editing as it might be useful for developing other applications.
\subsection{Spelling Correction of English Words}
Since the vendor collected data had borrowed English words the native transcribers committed some spelling mistakes especially in the English part of the sentence. So we extracted all the English words from the text data collected and manually mapped the words with the spelling corrected words and then replaced the mis-spelled words with the correct spellings.

\section{Data Preparation}
We have prepared the entire data language-wise with the following files that are necessary for ASR system building in ESPnet\cite{espnet_tool}.
\begin{itemize}
    \item text $=>$ ‘text’ file has utterance ID and the corresponding text. Format $:$$<$utterance\_ID$><$space$><$text corresponding to the utterance\_ID$>$
\item segments $=>$ ‘segments’ file has the start time and end time of a particular utterance in a particular speaker recording. 
Format $:$ $<$utterance\_ID$><$space$>$ $<$speake\\r\_ID$>$$<$space$><$start\_time$><$space$><$end\_time$>$
\item utt2spk $=>$ ‘utt2spk’ has the utterance\_ID information mapped to its corresponding speaker ID information. Format$:$ $<$utterance\_ID$><$space$><$speaker\_ID$>$
\item spk2utt $=>$ ‘spk2utt’ has speaker\_ID information mapped to utterances in that recording. Format$:$ $<$speaker\_ID$><$space$><$utterance\_ID$>$
\item wav.scp $=>$ ‘wav.scp’ file is nothing but a file having the recording\-ID and the path to where that particular recording is stored. Format$:$  $<$recording\_ID$><$space$>$ $<$path to the recording$>$
\end{itemize}

\section{Data Statistics}

\subsection{Number of Hours}

The prepared language-wise dataset was then split to train, valid and test tests. The number hours of training and validation data per language is presented in Table \ref{tab:hours}. All languages have a 5 hour test set.

\begin{table}[hbt!]
\centering
\begin{tabular}{c|c|c|c|c}
\hline
\hline
 Language  &Train &Valid & Test   &Total (Approx.)\\
 \hline
 \hline
 Assamese   &51 &5 &5 &61  \\
 Bengali   &375 &40 &5 &420 \\
 Gujarati  &176 &19 &5 &200\\  
Hindi  &316 &30 &5 &351 \\
Kannada  &83 &9 &5 &97 \\
Malayalam  &215 &25 &5 &245  \\
Marathi  &131 &14 &5 &150 \\
Odia  &82 &9 &5 &96  \\
Punjabi  &139 &15 &5 &159 \\
Tamil   &201 &20 &5 &226 \\ 
\hline

\end{tabular}

\caption{Total number of Hours Language-wise in SPRING-INX Data}
\label{tab:hours}
\end{table}

\subsection{Vocabulary Richness}

Vocabulary richness is an essential component of ASR data to build generalisable ASR models. In Table \ref{tab:tab3} we showcase the richness in vocabulary for different languages of SPRING-INX Train Data.

\begin{table}[hbt!]
\centering
\begin{tabular}{c|c|c}
\hline
\hline
 Language  & \#Native Words & \# Native Unique Words \\
 \hline
 \hline
 Assamese   &362012 &35950\\
 Bengali   &2603748 &111146 \\
 Gujarati  &1251027 &82971 \\  
Hindi  &2593053 &67839 \\
Kannada  &517095 &81256 \\
Malayalam  &1091017 &216729\\
Marathi  &903868 &99442\\
Odia  &610942 &39265  \\
Punjabi  &1171714 &47516 \\
Tamil   &1182484 &138986 \\ 
\hline
\end{tabular}

\caption{Vocabulary Richness of SPRING-INX Train Data}
\label{tab:tab3}
\end{table}

\subsection{Unique Speakers and Gender Balance}
The number of unique speakers per language along with the gender split are presented in Table \ref{tab:tab2}.
\begin{table}[hbt!]
\centering
\begin{tabular}{c|c|c}
\hline
\hline
 Language  & Male &Female\\
 \hline
 \hline
 Assamese   &180 &145 \\
 Bengali   &936 &797\\
 Gujarati  &476 &119\\  
Hindi  &764 &639 \\
Kannada  &200 &281 \\
Malayalam  &264 &188 \\
Marathi  &158 &224\\
Odia  &324 &402 \\
Punjabi  &257 &553 \\
Tamil   &160 &542\\ 
\hline
\end{tabular}

\caption{Gender Balance in SPRING-INX Data}
\label{tab:tab2}
\end{table}

\section{ESPnet Baselines}
For the researchers to benchmark their experiments, we have also made the ESPnet recipes available in the given URL: \textit{https://github.com/Speech-Lab-IITM/SPRING\_INX\_ESPnet\_R} \textit{ecipe}. The recipes are for building Transformer \cite{vaswani_transformer} models in joint CTC/attention framework \cite{hyb_CTCatt_shinji}.

\bibliographystyle{IEEEbib}
\bibliography{Paper}

\end{document}